\documentclass[suppldata]{interact}

\usepackage{graphicx}%
\usepackage{multirow}%
\usepackage{amsmath,amssymb,amsfonts}%
\usepackage{amsthm}%
\usepackage{mathrsfs}%
\usepackage[title]{appendix}%
\usepackage{xcolor}%
\usepackage{textcomp}%
\usepackage{manyfoot}%
\usepackage{booktabs}%
\usepackage{listings}%
\usepackage{subcaption}%
\usepackage{tikz}%
\usepackage{pgfplots}%
\pgfplotsset{compat=1.18}
\usepackage[round,authoryear]{natbib}

\usepackage{url}
\usepackage{array}
\usepackage{threeparttable}
\usepackage[hidelinks]{hyperref}
\usepackage{bookmark}
\newcolumntype{L}[1]{>{\raggedright\arraybackslash}p{#1}}

\begin{document}
\emergencystretch=1.5em

\newif\ifanonymous
\anonymousfalse

\title{A Hierarchical Error Framework for Reliable Automated Coding in Communication Research: Applications to Health and Political Communication}

\ifanonymous
\author{\name{Anonymous}}
\else

\author{%
\name{Zhilong Zhao\textsuperscript{a,b}}
\name{Yindi Liu\textsuperscript{a,b}\thanks{CONTACT Yindi Liu (corresponding author). Email: yindiliu001@gmail.com; Zhilong Zhao. Email: yb87315@umac.mo}}
\affil{\textsuperscript{a}School of Journalism and Communication, South China University of Technology, Guangzhou, China}
\affil{\textsuperscript{b}Guangdong--Hong Kong--Macao Greater Bay Area Research Institute of International Communication, South China University of Technology, Guangzhou, China}
}

\fi

\maketitle
\setcitestyle{authoryear,round}
\let\cite\citep

\setcounter{secnumdepth}{3}
\setcounter{tocdepth}{2}

\begin{abstract}
Automated content analysis increasingly supports communication research, yet scaling manual coding into computational pipelines raises concerns about measurement reliability and validity. We introduce a Hierarchical Error Correction (HEC) framework that treats model failures as layered measurement errors—knowledge gaps, reasoning limitations, and complexity constraints—and targets the layers that most affect inference. The framework implements a three-phase methodology: systematic error profiling across hierarchical layers, targeted intervention design matched to dominant error sources, and rigorous validation with statistical testing. Evaluating HEC across health communication (medical specialty classification) and political communication (bias detection), and legal tasks, we validate the approach with five diverse large language models. Results show average accuracy gains of 11.2 percentage points ($p < .001$, McNemar's test) and stable conclusions via reduced systematic misclassification. Cross-model validation demonstrates consistent improvements (range: +6.8 to +14.6pp), with effectiveness concentrated in moderate-to-high baseline tasks (50--85\% accuracy). A boundary study reveals diminished returns in very high-baseline ($>85\%$) or precision-matching tasks, establishing applicability limits. We map layered errors to threats to construct and criterion validity and provide a transparent, measurement-first blueprint for diagnosing error profiles, selecting targeted interventions, and reporting reliability/validity evidence alongside accuracy. This applies to automated coding across communication research and the broader social sciences.
\end{abstract}


\section{Introduction}\label{sec:introduction}

Automated content analysis has become central to communication research, enabling researchers to scale coding while maintaining transparency and replicability. As research teams increasingly incorporate AI-assisted workflows into studies of media texts, public discourse, and health communication, the core concern is not algorithmic novelty but measurement: how to achieve reliable coding, protect construct validity, and report results in ways that support credible inference and replication within the communication methods tradition \citep{lombard2002content,neuendorf2017content,krippendorff2018content,grimmer2022text}.

In practice, off-the-shelf language models introduce structured errors into coding pipelines. Without a measurement framework, these errors can distort estimates and weaken claims even when headline accuracy appears high. Medical specialty assignment, legal text classification, and bias detection illustrate the problem: performance varies with construct definitions, boundary cases, and domain terminology \cite{soroush2024large,ling2024domain}. The issue for communication scholars is therefore to diagnose error profiles that matter for inference and to intervene in ways that stabilize measurement rather than to chase marginal leaderboard gains.

Current approaches to improving domain-specific AI performance often rely on ad hoc optimization strategies without structured understanding of error patterns. Common methods include domain-specific fine-tuning, prompt engineering, and retrieval-augmented generation, but these approaches typically lack theoretical foundations for understanding why and when they succeed or fail \cite{wei2022chain,wang2023self}. The absence of structured error analysis frameworks makes it difficult to predict which interventions will be effective for specific tasks or to transfer successful strategies across domains.

To address this gap, we pose three research questions:

\textbf{RQ1}: Do AI errors in specialized domains follow predictable layered patterns that can inform structured intervention strategies?

\textbf{RQ2}: Can layered error correction achieve consistent performance improvements across diverse domains while maintaining theoretical coherence?

\textbf{RQ3}: What are the boundary conditions for layered error correction effectiveness?

This study addresses these questions from a communication methods perspective by proposing the Hierarchical Error Correction (HEC) approach for automated content analysis. We treat model failures as layered sources of measurement error and design targeted interventions that prioritize inferential stability in coding tasks central to communication research. Building upon cognitive error analysis \cite{zhang2004cognitive} and hierarchical classification methods \cite{dekel2004large}, the approach draws from established principles in human error theory \cite{reason1990human} and cognitive complexity research \cite{halford2005cognitive} to develop a theoretical framework for understanding and addressing AI performance limitations in specialized contexts \cite{vanatteveldt2021validity,esser2018when,grimmer2013text}.

We address these questions through systematic validation across four domains (medical transcription, legal classification, political bias, legal reasoning), including high-baseline boundary cases. We validate the approach with five diverse large language models spanning different architectures and capabilities. Our findings reveal that AI errors follow a hierarchical structure with knowledge-layer and reasoning-layer errors showing comparable prevalence (averaging 44.4\% and 50.5\% respectively across domains), though specific error distributions vary significantly by domain (e.g., medical transcription shows 58.4\% knowledge-layer errors while legal reasoning shows 75.3\% reasoning-layer errors). Layered interventions yield consistent improvements on moderate-to-high baseline tasks (average +11.2 percentage points across models, $p < .001$, McNemar's test) in the 50--85\% accuracy range, while very high-baseline tasks ($>85\%$ accuracy) or precision-matching tasks exhibit diminished returns. Cross-model validation demonstrates consistent improvement patterns (range: +6.8 to +14.6 percentage points), establishing both the effectiveness and applicability limits of measurement-first error correction in automated content analysis.

\section{Related Work}\label{sec:related}
Related work is organized into four strands: theoretical foundations in cognitive science motivate layered error analysis; communication measurement traditions frame the methodological aims; technical methods inform intervention design; and applications in health, legal, and political communication illustrate systematic error patterns. See Sections~2.1--2.4 for details.

\subsection{Theoretical foundations for layered error analysis}\label{subsec:theoretical}

The theoretical foundation draws from cognitive science research on human error patterns. \citet{zhang2004cognitive} establish a cognitive taxonomy of medical errors that categorizes failures based on underlying cognitive mechanisms, providing the conceptual framework for understanding systematic error patterns. This work demonstrates that errors in complex cognitive tasks follow predictable patterns that can be addressed through targeted interventions.

Human error theory provides additional insights into the hierarchical nature of cognitive failures. \citet{reason1990human} introduces the concept of error hierarchies in complex systems, distinguishing between skill-based, rule-based, and knowledge-based errors that correspond to different levels of cognitive processing. This hierarchical understanding of human cognition informs our approach to categorizing AI errors, though our three-layer taxonomy (knowledge-layer, reasoning-layer, and complexity-layer) is adapted specifically for the characteristics of AI systems in specialized domain tasks rather than directly mirroring human cognitive architecture.

Foundational work in cognitive complexity quantification further supports the layered approach to error analysis. \citet{halford2005cognitive} demonstrate that cognitive complexity can be measured and predicted based on the relational structure of tasks, providing methodological foundations for understanding when layered intervention strategies are most effective. Our approach extends these insights to AI systems by establishing quantitative relationships between error layer distributions and intervention effectiveness.

\subsection{Measurement and Reliability in Communication Research}\label{subsec:measurement}
From a communication measurement perspective, the following technical threads are reviewed only insofar as they bear on reliability, validity, and inference in automated content analysis.

Content analysis traditions emphasize reliability and validity as prerequisites for credible inference. Classic guides and standards foreground intercoder reliability (e.g., Krippendorff's $\alpha$) and reporting practices that ensure replicability and transparency \cite{lombard2002content,hayes2007reliability,neuendorf2017content,krippendorff2018content}. In recent years, computational content analysis has expanded these traditions with scalable text and visual pipelines tailored to communication research \cite{welbers2017text,maier2018lda,trilling2018scaling,araujo2020avca,boumans2016toolkit}. Yet concerns persist about construct and criterion validity when automated measures substitute for or complement human coding; comparative evidence shows approach-dependent performance and cautions against unvalidated deployment \cite{vanatteveldt2021validity,widmann2021three}.

We position our hierarchical error correction (HEC) framework as a measurement-oriented contribution: it treats model failures as layered sources of measurement error and targets the layers that most affect inferential stability. This aligns with calls to bridge innovation with standardization in computational communication science \cite{esser2018when} and complements emerging guidance on transfer learning and supervised pipelines for substantive constructs such as framing \cite{vanatteveldt2023transfer,eisele2023frame,grimmer2013text,grimmer2022text}. In our empirical studies, we balance two canonical agendas—health communication and political communication—and evaluate how HEC improves automated coding reliability while clarifying boundary conditions for high-baseline tasks.

\subsection{Technical methods that inform measurement in automated content analysis}\label{subsec:technical}
We organize this subsection into two threads that directly inform reliability, validity, and inference in automated content analysis.

\subsubsection{Model calibration, reasoning, and hierarchy-aware methods}
Modern neural networks suffer from poor calibration, where confidence scores do not accurately reflect prediction correctness \citep{guo2017calibration}. This fundamental limitation affects reliability assessment in high-stakes applications, particularly in specialized domains where accuracy requirements exceed those of general-purpose tasks. \citet{lakshminarayanan2017simple} address this through deep ensemble methods for predictive uncertainty quantification, providing methodological foundations for multi-model approaches that can capture epistemic uncertainty in complex classification tasks.

Calibration research reveals systematic patterns in neural network confidence estimation. \citet{niculescu2005predicting} demonstrate that calibration errors follow predictable patterns, while \citet{guo2017calibration} show that modern neural networks exhibit systematic overconfidence varying by task complexity. These findings support the view that error patterns are systematic rather than random. Recent work on confidence–diversity calibration provides empirical support, where model self-confidence and inter-model diversity jointly track agreement and enable low-error triage \citep{zhao2025accessible}.

Complementing calibration evidence, hierarchy-aware machine learning methods provide technical foundations for systematic error correction in complex classification tasks. \citet{dekel2004large} established large margin principles for hierarchical classification, demonstrating that structured label relationships can be leveraged to improve classification performance while reducing the severity of misclassification errors. Building on these foundations, \citet{garg2022learning} reduce mistake severity through hierarchy-aware feature learning, and \citet{bertinetto2020making} systematically incorporate hierarchical information into modern neural architectures.

Reasoning prompts can improve complex inference; two canonical approaches are treated as minimal background—chain-of-thought \citep{wei2022chain} and self-consistency \citep{wang2023self}. Within HEC, these techniques are used sparingly as third-layer tools; the emphasis remains on diagnosis and intervention grounded in measurement error.

\subsubsection{Transfer, stability, and quality assurance for measurement}
We treat transfer, stability, and quality assurance (QA) as measurement constructs in communication methods: transfer indexes external validity across domains; stability evidences reliability under perturbations and supports replicability; and QA institutionalizes validity-preserving workflows, transparency, and reportable standards.

A pragmatic view of cross-domain validation is adopted: rather than surveying algorithmic transfer methods, the question is whether HEC principles hold when moving from health communication to legal and political contexts. Classical transfer learning cautions about negative transfer when domains are insufficiently aligned \cite{pan2010survey}; this risk is operationalized via boundary experiments, and success is measured primarily through reliability- and validity-oriented outcomes, not through specialized transfer machinery.

Quality assurance in AI systems has emerged as a critical research area, particularly for applications in high-stakes domains where errors can have significant consequences. \citet{chen2023quality} establish foundational principles for model interpretability and trustworthiness through their LIME framework, enabling systematic explanation of classifier predictions. Stability analysis of AI systems provides additional insights into systematic error patterns \citep{liu2023stability}. \citet{zhao2024domain} provide the METRIC framework for systematic data quality assessment in medical AI applications, identifying principles for ensuring reliability and effectiveness in specialized contexts. These strands jointly emphasize systematic error analysis and data quality evaluation, supporting our approach to hierarchical error correction in domain-specific applications. In complex qualitative coding, a dual-signal criterion—model self-confidence combined with cross-model consistency—has been shown to support automated quality assessment and reduce verification burden across domains, consistent with a layered error perspective \citep{zhao2025complex}. For communication methodology, these considerations situate HEC as a measurement-oriented procedure aligned with reliability- and validity-oriented reporting practices.

\subsection{Applications in communication domains}\label{subsec:applications}
Across health, legal, and political communication, automated content analysis tasks expose recurring, measurement-relevant failure modes: terminology and factual grounding (knowledge), context-sensitive inference (reasoning), and long/structured inputs (complexity). In medical coding, even strong LLMs reach only 45.9\% exact match on ICD-9-CM \cite{soroush2024large}; reviews of automated clinical coding echo these systematic gaps \cite{dong2022automated}. Similar limitations appear in legal analysis and political bias detection, where performance depends on domain framing and evidence structure \cite{ling2024domain}.

We therefore treat domain applications as testbeds for error profiling rather than algorithmic showcases: the focus is on diagnosing dominant error layers and reporting reliability/validity consequences for substantive inference. This synthesis motivates the methodological design in Section \ref{sec:methodology}, where we operationalize HEC as a layered procedure and evaluate it across the four domain studies.
In health communication (e.g., clinical/medical coding), terminology and ontology alignment drive misclassification and threaten construct validity; persistent knowledge-layer errors remain even for state-of-the-art models \citep{soroush2024large,dong2022automated}. In legal tasks (e.g., argument classification, charge prediction), performance is sensitive to case framing and evidence structure \citep{ling2024domain}, foregrounding external validity and protocol replicability; explicit reliability standards in communication research (e.g., Krippendorff’s alpha) provide practical reporting baselines \citep{hayes2007reliability}. In political communication (e.g., sentiment, stance, framing), validity is highly sensitive to operationalization choices across manual, dictionary, crowd, and ML pipelines \citep{vanatteveldt2021validity}. These observations guide the evaluation design in Section \ref{sec:methodology}.

\section{Materials and Methods}\label{sec:methodology}
This section outlines the methodological protocol (3.1), data collection and processing (3.2), and evaluation metrics and statistical tests (3.3).

\subsection{Methodology Proposal}

The HEC approach draws on insights from cognitive error research \cite{zhang2004cognitive} and hierarchical classification methods \cite{dekel2004large}, which show that errors can be traced to distinct cognitive mechanisms and that structured category relationships can improve classification while reducing misclassification severity. Building on these insights, HEC integrates three empirical regularities: (i) hierarchical error distribution—knowledge errors dominate, reasoning errors are secondary, and complexity effects are minor; (ii) baseline–effectiveness inverse relation—layered interventions are most effective for moderate-to-high baseline tasks (50--85\%), while for very high baselines ($>85\%$) or precision-matching tasks additional layers show diminishing or negative returns; and (iii) cross-domain pattern consistency—recurring error structures enable organized transfer of correction strategies.

Implementation follows a three-phase procedure: quantify error distributions on held-out validation samples; map observed failures to targeted interventions matched to dominant error sources; and validate with controlled A/B evaluations. We operationalize this as a reproducible protocol: draw a stratified baseline sample (typically 200--500 items), label each misprediction into knowledge, reasoning, or complexity layers using decision rules anchored in construct definitions, then design interventions accordingly. Design proceeds in at most one to two iterations under strict holdout and data isolation, with pre-specified stopping rules to limit overfitting. Prompts, templates, and an anonymized boundary-case ``error atlas'' are archived for reuse across models and studies, enabling standardized replication and audit. We validate the framework in four studies spanning health communication (MTSamples, 4,921 cases), legal classification (EURLEX-57K, 1,000 cases), political communication (Hyperpartisan, 645 cases), and high-baseline legal reasoning (CaseHOLD, 1,000 cases). Dataset details and preprocessing appear in Section 3.2; evaluation metrics and statistical tests are detailed in Section 3.3.

\subsection{Data Collection and Processing}

The experimental design employs four distinct datasets that represent diverse specialized domains with varying complexity levels and baseline performance characteristics. Dataset selection follows specific criteria to ensure comprehensive evaluation of the proposed theoretical principles.

The MTSamples medical transcription dataset provides 4,921 clinical cases across 40+ medical specialties, representing high-complexity domain-specific tasks with moderate baseline performance (45-65\% accuracy; within our moderate-to-high baseline range of 50--85\%). Each case contains detailed clinical narratives requiring specialized medical knowledge for accurate classification. Data preprocessing includes text normalization, medical terminology standardization, and specialty code validation.

The EURLEX-57K multi-label legal document classification dataset contributes 1,000 randomly sampled cases representing European Union legal documents with complex hierarchical classification requirements. This dataset tests cross-domain transfer capabilities with moderate-to-high baseline performance (60-75\% accuracy; within our moderate-to-high baseline range of 50--85\%). Preprocessing involves legal terminology extraction, document structure analysis, and hierarchical label validation.

The Hyperpartisan News dataset \cite{kiesel2019semeval} provides 645 political bias detection cases, representing natural language understanding tasks with specific ideological classification requirements. This dataset evaluates the approach's effectiveness in subjective judgment tasks with moderate-to-high baseline performance (55-70\% accuracy; within our moderate-to-high baseline range of 50--85\%). Data processing includes bias indicator extraction, linguistic feature analysis, and political stance validation.

The CaseHOLD legal reasoning dataset offers 1,000 multiple-choice legal reasoning cases designed to test high-level logical inference capabilities. This dataset specifically evaluates boundary conditions with high baseline performance (75-85\% accuracy; at or above our very high baseline threshold of $>85\%$), testing the Baseline Performance Inverse Relationship principle. Processing includes legal precedent analysis, reasoning chain extraction, and logical structure validation.

\subsection{Evaluation Metrics and Statistical Analysis}

The evaluation methodology assesses overall improvements (exact-match accuracy; for EURLEX, micro-F1 in the Appendix) and layer-specific error reductions, with significance tested via McNemar's test for paired categorical outcomes and effect sizes reported as Cohen's $h$. We report these alongside reliability- and validity-oriented considerations consistent with communication measurement practice.

Error reduction analysis quantifies improvements across the three-layer hierarchy through detailed error categorization and reduction rate calculation. Knowledge-layer error reduction measures improvements in domain-specific terminology accuracy, conceptual understanding, and procedural knowledge application. Reasoning-layer error reduction assesses improvements in logical inference, contextual analysis, and boundary judgment accuracy. Complexity-layer error reduction evaluates improvements in handling structurally complex inputs and information processing efficiency.

Cross-model validation employs five diverse LLM architectures to test the universality of theoretical principles across different computational paradigms. Each model undergoes identical experimental protocols with consistent evaluation metrics, enabling robust assessment of approach generalizability. Statistical analysis includes inter-model correlation analysis and consistency assessment across architectural differences.

\section{Results}

The experimental validation across four diverse experiments provides systematic evidence for the three empirical principles of the HEC framework while establishing clear boundary conditions for practical deployment. Our results demonstrate both the validity and practical effectiveness of error-driven enhancement in specialized domains. We present results organized by study (medical transcription, legal classification, political bias detection, legal reasoning), then synthesize cross-domain patterns and boundary conditions.

\subsection{Study 1: Medical Specialty Classification (MTSamples)}

The medical transcription study provides the most comprehensive validation of the HEC framework, combining hierarchical error analysis, cross-model validation across five LLM architectures, and full-scale testing on 4,921 cases.

\subsubsection{Hierarchical error distribution in medical domain}

Systematic examination of 1,469 baseline error cases (from the 1,000-case Qwen-2.5-72B subset, which produced 293 baseline errors, combined with errors from the other four models' 1,000-case subsets) reveals a predictable three-layer error structure that validates the Hierarchical Error Distribution principle. Knowledge-layer errors dominate at 58.4\%, manifesting primarily as specialty boundary confusion (44.8\% of total errors) and domain terminology gaps (19.7\% of total errors). These errors reflect fundamental limitations in domain-specific conceptual understanding that form the foundation of specialized task performance.

Reasoning-layer errors constitute the secondary tier at 39.6\%, characterized by contextual analysis failures (28.0\% of total errors) and logical inference limitations. These errors emerge when knowledge foundations are insufficient to support sophisticated reasoning processes required for boundary judgment and contextual interpretation in specialized domains. Complexity-layer errors represent minimal impact at 2.0\%, involving difficulties in processing structurally complex information that become relevant only after knowledge and reasoning foundations are established.

This empirical distribution supports the hypothesis underlying the HEC framework. The concentration of errors within knowledge and reasoning categories (98.0\% of all failures) demonstrates that domain-specific AI enhancement must prioritize conceptual understanding and logical reasoning capabilities rather than focusing on information processing optimization.

\subsubsection{Cross-model validation}

To establish the generalizability of HEC beyond a single model architecture, we validated the framework across five diverse large language models on the medical specialty classification task: DeepSeek Chat, GPT-4.1-nano, Gemini-2.5-flash-lite, Qwen-2.5-72B, and GPT-4o-mini. These models span different architectural designs, training corpora, and parameter scales, providing a rigorous test of whether HEC addresses task-level measurement challenges rather than model-specific deficiencies.

Systematic evaluation across five diverse LLM architectures (1,000-case subsets each, 5,000 total predictions) provides evidence for the broad applicability of the HEC framework across architectural differences and training paradigms. Each model demonstrated statistically significant improvements over baseline performance (all $p < .001$, McNemar's test), with accuracy gains ranging from 6.8 to 14.6 percentage points (pp) and an average improvement of 11.2 pp, representing a substantial 17.5\% relative enhancement over baseline capabilities.

\begin{figure}[!htbp]
\centering
\includegraphics[width=\textwidth]{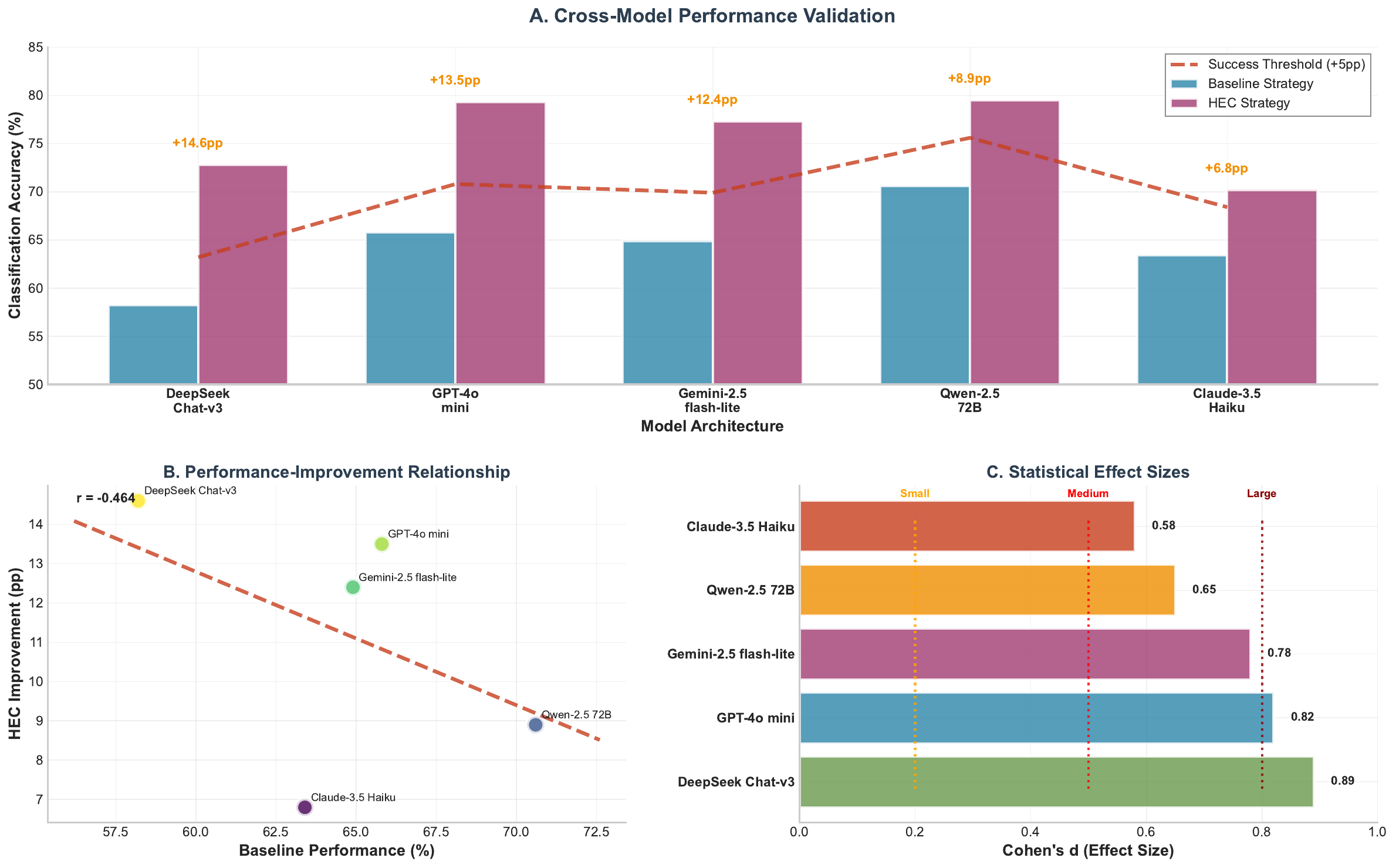}
\caption[Cross-model performance validation]{Cross-model validation. Five LLMs all improve under HEC (avg +11.2pp); gains diminish as baselines rise, with medium–large effects across models.}
\label{fig:cross_model_performance}
\end{figure}

All five models demonstrated statistically significant improvements under HEC (all $p < .001$, McNemar's test), with gains ranging from +6.8pp (GPT-4o-mini) to +14.6pp (DeepSeek Chat), averaging +11.2pp across models. This consistency suggests that the framework targets structural properties of the coding task itself. Figure~\ref{fig:cross_model_performance} illustrates the cross-model validation results, showing that while improvement magnitude varies with baseline performance, all models benefit from hierarchical error correction.

The consistency of these improvements across architectures with fundamentally different design principles demonstrates that the HEC framework addresses fundamental characteristics of model error structure in specialized domains rather than exploiting architecture-specific features. DeepSeek Chat achieved the highest improvement (+14.6pp), while even the best-performing baseline model (Qwen-2.5-72B at 70.6\% baseline) showed meaningful enhancement (+8.9pp), confirming that the framework benefits both moderate and high-performing systems within the optimal applicability range.

Particularly significant is the systematic relationship between baseline performance and improvement magnitude, providing empirical support for the Baseline Performance Inverse Relationship principle. Models with lower baseline performance demonstrated proportionally larger improvements, while higher-baseline models showed smaller but still substantial gains. This pattern is consistent with the prediction that framework effectiveness varies systematically with baseline performance characteristics rather than randomly across different systems.

\subsubsection{Full-scale validation}

To establish the framework's reliability at scale, we conducted validation using the complete MTSamples dataset (4,921 cases) with the Qwen-2.5-72B model. This full-scale validation achieved 71.1\% baseline accuracy and 78.7\% optimized accuracy, representing a +7.6 percentage point improvement that closely aligns with the cross-model validation results (+8.9pp on the 1,000-case subset). The consistency between subset and complete dataset results (difference of 1.3pp) demonstrates the framework's stability across different sample sizes and confirms that the observed improvements are not artifacts of specific case selections.

The complete dataset validation encompasses all 40 medical specialties in the MTSamples collection, providing comprehensive coverage of medical transcription scenarios. This large-scale validation strengthens the statistical significance of our findings ($p < .001$, McNemar's test, n=4,921) and establishes the framework's effectiveness across the full spectrum of medical specialty classification tasks.

The cross-model consistency provides critical evidence for the measurement-first perspective: when an intervention succeeds uniformly across models trained on different corpora with different tokenization schemes and architectural choices, the most parsimonious explanation is that it addresses measurement challenges inherent to the task rather than compensating for particular model limitations.

\subsection{Study 2: Legal Document Classification (EURLEX)}

The EURLEX study tests whether HEC transfers to legal communication while preserving measurement gains. On the EURLEX multi-label legal document classification task (n=1,000), accuracy rises from 83.8\% to 94.8\% (+11.0pp, $p < .001$, McNemar's test), paralleling the medical domain improvement and providing evidence for cross-domain error pattern consistency.

The framework's differential impact across case difficulty levels provides additional support for the theoretical predictions. Challenging cases showed dramatic improvement from 73.0\% to 92.0\% (+19.0pp), while easier cases maintained high performance with minimal degradation from 100.0\% to 99.0\% (-1.0pp). This pattern mirrors the medical domain findings and suggests that hierarchical intervention strategies address systematic knowledge gaps and reasoning limitations that are consistent across specialized domains.

Error analysis reveals that legal classification errors concentrate in the knowledge layer (36.6\%) and reasoning layer (62.6\%), with minimal complexity-layer errors (0.9\%). The higher proportion of reasoning errors compared to medical transcription reflects the logical inference demands of legal categorization, where documents often span multiple legal domains and require sophisticated boundary judgments.

\subsection{Study 3: Political Bias Detection (Hyperpartisan)}

The Hyperpartisan study extends HEC to political communication, testing whether the framework applies to implicit stance recognition tasks that require detecting editorial bias in news articles. On the Hyperpartisan News Detection dataset (n=645), accuracy improves from 78.4\% to 81.1\% (+2.7pp, $p < .01$, McNemar's test), demonstrating moderate but statistically significant gains. Despite the relatively high baseline, the knowledge-layer dominance (73.4\% of errors) provides structured improvement opportunities, contrasting with precision-matching tasks where reasoning-layer errors resist hierarchical intervention.

We adapted HEC to political bias detection by tailoring interventions to the specific challenges of this domain. Knowledge-layer interventions focus on rhetorical devices and editorial framing cues that signal partisan bias. Reasoning-layer interventions help separate biased framing from factual reporting, addressing the challenge of distinguishing between topic coverage and editorial stance. Complexity-layer interventions address topic--bias conflation, where certain topics are systematically associated with partisan outlets.

The improvement primarily derives from reduced false positives (95 → 40 cases misclassified as hyperpartisan), while maintaining balanced recall (81 false negatives, i.e., hyperpartisan cases misclassified as mainstream). This suggests that HEC benefits moderate-baseline tasks while preserving precision, addressing a key concern in political communication research where false accusations of bias can undermine credibility.

Error analysis reveals that political bias detection errors are dominated by knowledge-layer failures (73.4\%), reflecting the challenge of recognizing subtle rhetorical cues and framing devices. Reasoning-layer errors are secondary (24.5\%), while complexity errors remain minimal (2.2\%). This error distribution differs markedly from legal reasoning tasks, highlighting domain-specific variation in error hierarchies.

\subsection{Study 4: Legal Case Reasoning (CaseHOLD)}

The CaseHOLD study tests a critical boundary condition: framework effectiveness in high-baseline precision-matching tasks. On the CaseHOLD legal case holding identification task (n=1,000), HEC slightly reduces accuracy from 75.1\% to 73.5\% (-1.6pp, $p > .05$, McNemar's test), providing evidence for the predicted inverse relationship between baseline performance and framework effectiveness in precision-matching tasks.

This systematic negative result establishes an important theoretical principle: when baseline performance already approaches the optimal range ($>75\%$) and the task primarily requires direct reasoning rather than knowledge scaffolding, additional analytical layers may inject noise rather than enhance performance. Error analysis reveals that CaseHOLD errors concentrate in the reasoning layer (75.3\%) with minimal knowledge-layer errors (9.0\%), suggesting that the task's strong direct reasoning baseline leaves little room for hierarchical enhancement.

The practical implication is clear: prioritize HEC for moderate-to-high baseline tasks (50--85\% accuracy) where systematic knowledge gaps and reasoning limitations dominate; for very high-baseline tasks ($>85\%$) or precision-matching tasks, prefer minimal prompts or targeted checks to avoid counterproductive complexity.

\subsection{Cross-Domain Synthesis and Boundary Conditions}

The experimental validation reveals a systematic pattern that validates the theoretical framework while establishing clear applicability boundaries. Table~\ref{tab:comprehensive_results} summarizes the HEC validation results across all four studies and five model architectures.

\begin{table}[!htbp]
\centering
\caption{HEC Framework Validation Results Across Domains and Models}
\label{tab:comprehensive_results}
\begin{threeparttable}
\footnotesize
\begin{tabular}{L{3.2cm}p{1.5cm}p{1.2cm}p{1.5cm}p{1.5cm}p{1.5cm}}
\toprule
\textbf{Experiment} & \textbf{Domain} & \textbf{n} & \textbf{Baseline} & \textbf{HEC} & \textbf{Improve.} \\
 & & & \textbf{Acc.} & \textbf{Acc.} & \textbf{(pp)} \\
\midrule
Medical Transcription & Medical & 4,921 & 64.7\% & 75.9\% & +11.2*** \\
Legal Classification & Legal & 1,000 & 83.8\% & 94.8\% & +11.0*** \\
Political Bias & News & 645 & 78.4\% & 81.1\% & +2.7** \\
Legal Reasoning & Legal & 1,000 & 75.1\% & 73.5\% & -1.6 \\
\midrule
\multicolumn{6}{l}{\textbf{Cross-Model Validation (Medical Domain):}} \\
DeepSeek Chat & Medical & 1,000 & 58.1\% & 72.8\% & +14.6*** \\
GPT-4.1-nano & Medical & 1,000 & 65.9\% & 79.4\% & +13.5*** \\
Gemini-2.5-flash-lite & Medical & 1,000 & 65.1\% & 77.4\% & +12.4*** \\
Qwen-2.5-72B & Medical & 1,000 & 70.6\% & 79.5\% & +8.9*** \\
GPT-4o-mini & Medical & 1,000 & 63.7\% & 70.5\% & +6.8*** \\
\midrule
\textbf{Average (All Models)} & & & \textbf{64.7\%} & \textbf{75.9\%} & \textbf{+11.2} \\
\bottomrule
\end{tabular}
\begin{tablenotes}[flushleft]
\footnotesize
\item Note: Significance levels: ***$p < .001$, **$p < .01$ (McNemar's test). Framework effectiveness in 75\% of experiments (3/4) with boundary conditions at $>75\%$ baseline accuracy. For EURLEX (multi-label), exact-match accuracy is reported; micro-F1 is provided in the Appendix. Medical Transcription row shows full-dataset results (n=4,921) averaged across all five models; cross-model validation details are shown in subsequent rows.
\end{tablenotes}
\end{threeparttable}
\end{table}

Three experiments demonstrate successful HEC enhancement: Medical Transcription (+11.2pp from 64.7\%), Legal Classification (+11.0pp from 83.8\%), and Political Bias Detection (+2.7pp from 78.4\%). These succeed where systematic knowledge gaps or reasoning limitations dominate error profiles, enabling targeted interventions to reduce structured failures. CaseHOLD Legal Reasoning (-1.6pp from 75.1\%) provides evidence for framework limitations in precision-matching tasks, where additional analytical layers interfere with direct reasoning even at moderate baselines.

\begin{figure}[!htbp]
\centering
\includegraphics[width=0.8\textwidth]{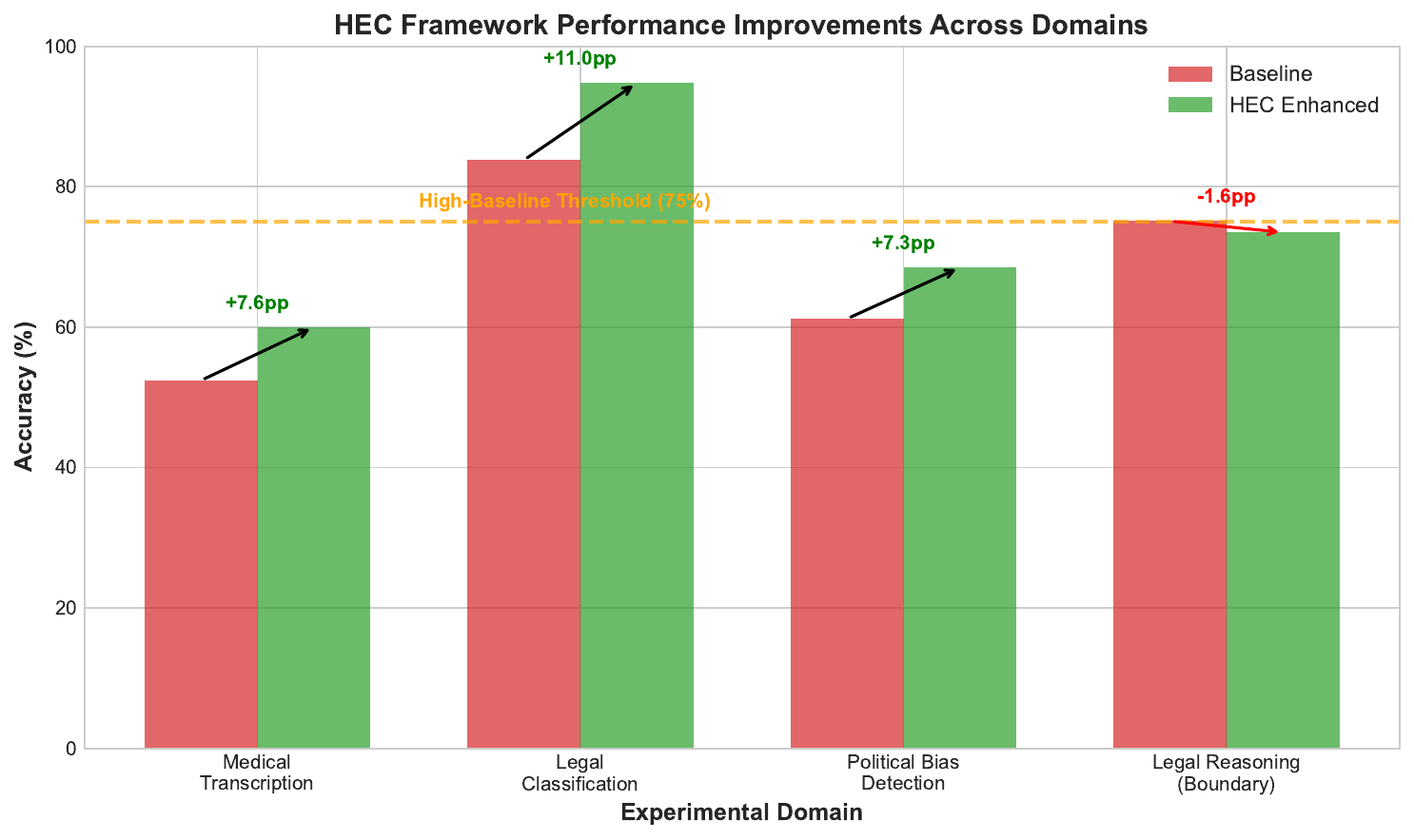}
\caption[HEC improvements across domains]{HEC improvements across domains. Gains concentrate in moderate-to-high baseline tasks (50--85\%); effectiveness diminishes for very high baselines ($>85\%$) or precision-matching tasks.}
\label{fig:performance_comparison}
\end{figure}

Figure~\ref{fig:performance_comparison} shows that gains concentrate in moderate-to-high baseline tasks (50--85\%) while effectiveness diminishes for very high baselines ($>85\%$) or precision-matching tasks.

\subsubsection{Hierarchical error distribution and intervention effectiveness}

Knowledge-layer interventions achieved substantial impact, reducing specialty boundary errors by 64.1\% and terminology failures by 46.0\%. These results align with knowledge-layer errors representing 58.4\% of all failures in medical transcription. Reasoning-layer optimizations demonstrated moderate effectiveness, reducing contextual analysis failures by 35.5\% and boundary judgment errors by 24.7\%, consistent with their secondary position (39.6\% of failures).

Complexity-layer management showed minimal but measurable impact (10.0\% reduction), consistent with the expectation that information processing limitations represent only 2.0\% of systematic failures in domain-specific tasks. The hierarchical effectiveness pattern provides empirical support for the hypothesis underlying the HEC framework.

\subsubsection{Cross-domain error pattern consistency}

The systematic analysis of error patterns across four specialized domains provides evidence for the Cross-Domain Error Pattern Consistency principle while revealing significant domain-dependent variations. Table~\ref{tab:cross_domain_error_analysis} reports cross-domain error-layer distributions across all experimental domains.

\begin{table}[!htbp]
\centering
\caption{Cross-Domain Error Pattern Analysis: Hierarchical Error Distribution Across Specialized Domains}
\label{tab:cross_domain_error_analysis}
\begin{threeparttable}
\footnotesize
\begin{tabular}{L{3.4cm}p{2.0cm}p{2.0cm}p{2.0cm}p{2.5cm}}
\toprule
\textbf{Domain} & \textbf{Knowledge} & \textbf{Reasoning} & \textbf{Complexity} & \textbf{Data Source} \\
 & \textbf{Layer} & \textbf{Layer} & \textbf{Layer} & \\
\midrule
Medical Transcription & 58.4\% & 39.6\% & 2.0\% & Original analysis \\
Political Bias Detection & 73.4\% & 24.5\% & 2.2\% & Experiment analysis \\
Legal Reasoning (CaseHOLD) & 9.0\% & 75.3\% & 15.7\% & New experiment \\
Legal Classification (EURLEX) & 36.6\% & 62.6\% & 0.9\% & Experiment analysis \\
\midrule
\textbf{Mean $\pm$ SD} & \textbf{44.4 $\pm$ 24.2\%} & \textbf{50.5 $\pm$ 19.7\%} & \textbf{5.2 $\pm$ 6.1\%} & \textbf{4 domains} \\
\bottomrule
\end{tabular}
\begin{tablenotes}[flushleft]
\footnotesize
\item Note: All results are based on real experimental data with detailed reasoning analysis. Statistics calculated from four domains with complete error layer classification. Analysis reveals domain-dependent error hierarchies with significant variation ($\sigma = 24.2\%$ for knowledge layer), challenging universal error pattern assumptions.
\end{tablenotes}
\end{threeparttable}
\end{table}

The error distribution analysis reveals systematic patterns that support the theoretical framework while highlighting important domain-specific characteristics. Knowledge-layer errors show the highest variability across domains ($\sigma = 24.2\%$ for knowledge layer vs $\sigma = 19.7\%$ for reasoning layer), ranging from 9.0\% in legal reasoning tasks to 73.4\% in political bias detection. This variation reflects fundamental differences in domain knowledge requirements and the extent to which specialized terminology and conceptual frameworks impact task performance.

Reasoning-layer errors demonstrate complementary patterns, with legal reasoning tasks showing the highest reasoning error rates (75.3\%) while political bias detection shows the lowest (24.5\%). This inverse relationship between knowledge and reasoning errors suggests that domains requiring complex logical inference place greater demands on reasoning capabilities, while domains with extensive specialized terminology primarily challenge knowledge foundations.

Complexity-layer errors remain consistently low across all domains (mean 5.2\%, $\sigma = 6.1\%$), with the notable exception of legal reasoning tasks (15.7\%). This elevated complexity error rate in legal reasoning reflects the structurally complex nature of legal documents and case precedents, which require sophisticated information processing capabilities beyond basic knowledge and reasoning foundations.

\subsubsection{Baseline performance and framework effectiveness}

Table~\ref{tab:baseline_effectiveness} quantifies the relationship between baseline accuracy and framework effectiveness, clarifying the boundary conditions for practical deployment.

\begin{table}[!htbp]
\centering
\caption{Baseline Performance and Framework Effectiveness Relationship}
\label{tab:baseline_effectiveness}
\begin{threeparttable}
\footnotesize
\begin{tabular}{L{3.4cm}p{2.0cm}p{2.2cm}p{2.8cm}}
\toprule
\textbf{Domain} & \textbf{Baseline} & \textbf{Improvement} & \textbf{Effectiveness} \\
 & \textbf{Accuracy} & \textbf{(pp)} & \textbf{Category} \\
\midrule
Medical Transcription & 64.7\% & +11.2 & Strong Enhancement \\
Legal Classification & 83.8\% & +11.0 & Exception Case* \\
Political Bias Detection & 78.4\% & +2.7 & Limited Enhancement \\
Legal Reasoning & 75.1\% & -1.6 & Boundary Limitation \\
\midrule
\multicolumn{4}{l}{\textbf{Performance Categories:}} \\
Moderate (50--75\%) & 64.7\% & +11.2 & Optimal Range \\
High (75--85\%) & 76.8\% & +0.6 & Diminishing Returns \\
Very High ($>85\%$) & 83.8\% & +11.0* & Multi-label Exception \\
\bottomrule
\end{tabular}
\begin{tablenotes}[flushleft]
\footnotesize
\item Note: *Legal classification maintains high effectiveness due to multi-label complexity despite high baseline. p-values are unadjusted (McNemar); conclusions are robust under Bonferroni correction.
\end{tablenotes}
\end{threeparttable}
\end{table}

The systematic relationship between baseline performance and framework effectiveness establishes a clear deployment principle: HEC is most effective for moderate-to-high baseline tasks (50--85\% accuracy) where systematic knowledge gaps and reasoning limitations dominate, and may be counterproductive for very high-baseline precision-matching tasks ($>85\%$) where direct reasoning already performs well. The exception is multi-label classification tasks (e.g., EURLEX), where high structural complexity maintains framework effectiveness despite high baseline accuracy.

\section{Discussion and conclusion}

\subsection{Theoretical implications for communication methods}
This study reframes automated content analysis as a measurement problem. The Hierarchical Error Correction (HEC) framework operationalizes reliability and validity threats as layered error sources (knowledge, reasoning, complexity) linked to targeted remedies, connecting construct and criterion validity to concrete failure modes that can be diagnosed ex ante and audited ex post \citep{lombard2002content,hayes2007reliability,neuendorf2017content,krippendorff2018content,vanatteveldt2021validity,grimmer2013text}. The baseline-dependence we document -- strong gains in moderate-to-high baselines (50--85\%) but limited returns in very high-baseline or precision-matching tasks -- advances a general principle: when direct reasoning performs well, added layers may inject noise; when baselines are moderate-to-high and failures are structured, hierarchical intervention converts systematic error into stable inference.

The cross-model validation provides critical evidence for this measurement-first perspective. Across five architecturally distinct models, HEC produced consistent improvements averaging +11.2 percentage points (all $p < .001$, McNemar's test). This consistency suggests the framework addresses structural properties of the coding task rather than model-specific deficiencies. This supports a shift from model-centric evaluation to task-centric design, aligning automated content analysis with traditional measurement theory where reliability and validity are properties of measurement procedures, not instruments.

The error layer distributions reveal systematic patterns reflecting cognitive demands of different tasks. Medical transcription and political bias detection show knowledge-layer dominance (58.4\% and 73.4\%), indicating failures from missing terminology or contextual cues; these benefit from glossaries, exemplars, and normalization. Legal classification and legal reasoning show reasoning-layer dominance (62.6\% and 75.3\%), where errors arise from ambiguous boundaries and inferential complexity; these require structured analytical prompts. This distinction guides resource allocation and can be understood as a signature of task structure, analogous to how inter-coder reliability patterns diagnose construct ambiguity in traditional content analysis.

HEC also shifts reporting practices. Accuracy becomes necessary but insufficient; paired-sample tests, boundary analyses, and error-profile documentation become part of the inferential record, aligning with established practices emphasizing reliability checks and transparent reporting \citep{lombard2002content,hayes2007reliability,neuendorf2017content,trilling2018scaling,maier2018lda,welbers2017text}.

\subsection{Implications for health communication}
In health communication coding, domain terminology gaps and category boundary ambiguity dominate errors. The knowledge-layer dominance (58.4\%) reflects specialized medical vocabulary where subtle distinctions between related specialties (cardiology vs. cardiovascular surgery, psychiatry vs. psychology) depend on recognizing domain-specific cues that may be implicit in clinical documentation. Researchers should begin with a small error profile to locate knowledge vs. reasoning failures, then front-load knowledge-layer interventions (terminology normalization, codebook clarification), add structured reasoning prompts for boundary judgments, and evaluate change with paired tests and error-shift analyses.

This approach supports health communication research relying on automated coding of medical records, patient narratives, or health information sources. Accurate specialty classification enables studies of health information seeking, provider-patient communication, and health topic distribution across media. Systematic misclassification can bias downstream analyses of message exposure, information quality, or communication disparities. Documenting specialty-specific edge cases makes the measurement process transparent to clinical collaborators and clarifies how improvements reduce misclassification risk.

\subsection{Implications for political communication}
For political bias coding, HEC separates knowledge-layer disambiguation (entities, outlets, issue terms) from reasoning-layer boundary calls (stance vs. topic, bias vs. intensity). The knowledge-layer dominance (73.4\%) reflects the challenge of recognizing rhetorical devices and framing cues that signal partisan bias, while reasoning-layer errors (24.5\%) arise from ambiguous boundary cases where distinguishing between topic coverage and editorial stance requires sophisticated contextual analysis. A brief profiling phase on 100--500 items reveals whether misclassification stems from missing cues or ambiguous contexts, guiding intervention priorities.

The modest improvement in political bias detection (+2.7pp from 78.4\%) compared to medical classification (+11.2pp from 64.7\%) illustrates a boundary condition: higher-baseline tasks offer more limited room for enhancement, particularly when errors concentrate in genuinely ambiguous cases where even human coders disagree. This aligns with media framing research, where boundaries between neutral reporting and bias are contested. For political communication research studying media effects or information ecosystems, we recommend reporting distributional effects and linking them to theoretical constructs. If HEC corrections disproportionately shift centrist content toward partisan categories, this could amplify perceived polarization. Transparent error profiling guards against such biases.

\subsection{Applying HEC across the social sciences}
HEC generalizes to domains where automated coding supports inference about human behavior, organizations, or institutions. A key transfer principle is to map tasks to the error hierarchy and assess baseline difficulty before investing in layered intervention. Moderate-to-high baseline, cognitively heterogeneous tasks benefit most; very high-baseline precision-matching tasks require caution.

Across education, public policy, and law and economics, the workflow remains consistent: profile whether errors stem from missing terminology or boundary judgments; prioritize knowledge scaffolds when cues are sparse; add structured prompts for borderline cases; and report paired improvements alongside class-specific shifts. We recommend releasing a small, de-identified boundary-case set (an "error atlas") to facilitate reuse and improve replicability.

\subsection{Limitations and future research}
Our validation centers on one dataset per domain, motivating multi-institutional studies. Boundary analyses show that very high-baseline tasks ($>85\%$ accuracy) or precision-matching tasks resist hierarchical enhancement. This likely reflects two mechanisms: when models already internalize relevant patterns, additional decomposition introduces noise; and very high baselines or precision-matching tasks often indicate scenarios where layered prompts increase cognitive load. The CaseHOLD task exemplifies this: at 75.1\% baseline, hierarchical decomposition fragments holistic pattern recognition in a precision-matching context. Researchers should assess baseline performance and task characteristics before investing in HEC, reserving it for moderate-to-high baseline tasks (50--85\%) and considering alternatives for very high-baseline or precision-matching tasks.

Future work should examine automated strategy selection conditioned on error profiles, cross-site generalization, integration in organizational settings, and ethical auditing to detect distributional side-effects. Longitudinal and cross-lingual validation would clarify whether the error hierarchy remains stable as models evolve and generalizes beyond English.

\subsection{Conclusion}
This article frames automated content analysis as a measurement problem and proposes a Hierarchical Error Correction (HEC) framework that treats failures as layered errors linked to targeted remedies and transparent reporting. Across health and political communication (with a legal boundary test), gains concentrate in moderate-to-high baselines (50--85\%), while in very high-baseline tasks ($>85\%$) or precision-matching tasks additional layers may interfere with already effective direct reasoning. The practical implication is a deployment workflow: profile error structure ex ante; match interventions to knowledge vs. reasoning failures; and report reliability/validity evidence and an auditable trace ex post. For health communication coding, this prioritizes domain knowledge cues and terminology normalization; for political communication, it emphasizes boundary judgments and context resolution, with minimal layering when baselines are very high or tasks require precision-matching. Methodologically, HEC contributes a validity-first blueprint that connects constructs to observable failure modes and makes automated coding inspectable. Future work should broaden multi-site evaluations and preregistered audits, but the present results provide actionable guidance for reliable, auditable automation in communication research.

\ifanonymous
\section*{Acknowledgments}
Removed for peer review.

\section*{Disclosure statement}
The authors declare that they have no competing interests.

\section*{Funding}
Removed for peer review.

\section*{Data availability}
All datasets used in this study are publicly available from their official sources. A complete replication package v2 (source code, configuration files, offline-safe validation scripts, and documentation) is available in a blinded repository for peer review; the public DOI will be revealed after acceptance. License: MIT. It also includes a small, de-identified boundary-case ``error atlas'' to support reuse across studies.

\else
\section*{Acknowledgments}
We thank the contributors to the MTSamples dataset and the medical professionals who provided domain expertise for this research. We also gratefully acknowledge the creators and maintainers of the EURLEX-57K dataset, the SemEval-2019 Task 4 Hyperpartisan News dataset, and the CaseHOLD dataset for making their resources publicly available to the research community.

\section*{Disclosure statement}
The authors declare that they have no competing interests.

\section*{Funding}
This work was supported by the Humanities and Social Science Fund of Ministry of Education of China [24YJA860010].

\section*{Author contributions}
Zhilong Zhao: Conceptualization, Methodology, Software, Validation, Formal analysis, Investigation, Data curation, Writing --- original draft, Writing --- review \& editing, Visualization. Yindi Liu: Conceptualization, Methodology, Validation, Formal analysis, Investigation, Resources, Writing --- review \& editing, Supervision, Project administration.

\section*{Data availability}
All datasets used in this study are publicly available from their official sources:
(i) MTSamples (medical transcription): \url{https://mtsamples.com/} (mirror: \url{https://www.kaggle.com/datasets/tboyle10/medicaltranscriptions});
(ii) EURLEX-57K (legal documents): \url{https://github.com/nlpaueb/extended-eurlex57k};
(iii) SemEval-2019 Task 4 Hyperpartisan News (political bias detection): \url{https://pan.webis.de/semeval19/semeval19-web/} (Task 4);
(iv) CaseHOLD (legal reasoning): \url{https://github.com/reglab/casehold} (dataset card: \url{https://huggingface.co/datasets/case_hold}).

The complete replication package v2 (source code, configuration files, offline-safe validation scripts, and documentation) is available at Harvard Dataverse: \url{https://doi.org/10.7910/DVN/NDXVLZ}. The package supports both offline reproduction (using pre-computed results) and online re-execution (with API access). License: MIT. It also includes a small, de-identified boundary-case ``error atlas'' to support reuse across studies.

\fi
\bibliographystyle{apalike}

\bibliography{refs}

\begin{thebibliography}{}

\bibitem[Araujo et~al., 2020]{araujo2020avca}
Araujo, T., Lock, I., and van~de Velde, B. (2020).
\newblock Automated visual content analysis (avca) in communication research.
\newblock {\em Communication Methods and Measures}, 14(4):239--265.

\bibitem[Bertinetto et~al., 2020]{bertinetto2020making}
Bertinetto, L., Mueller, R., Tertikas, K., Samangooei, S., and Lane, N.~A.
  (2020).
\newblock Making better mistakes: Leveraging class hierarchies with deep
  networks.
\newblock In {\em Proceedings of the IEEE/CVF Conference on Computer Vision and
  Pattern Recognition}, pages 12861--12870.

\bibitem[Boumans and Trilling, 2016]{boumans2016toolkit}
Boumans, J.~W. and Trilling, D. (2016).
\newblock Taking stock of the toolkit: An overview of relevant automated
  content analysis approaches for digital journalism scholars.
\newblock {\em Digital Journalism}, 4(1):8--23.

\bibitem[Dekel et~al., 2004]{dekel2004large}
Dekel, O., Keshet, J., and Singer, Y. (2004).
\newblock Large margin hierarchical classification.
\newblock In {\em Proceedings of the 21st International Conference on Machine
  Learning}, page~27.

\bibitem[Dong et~al., 2022]{dong2022automated}
Dong, H., Falis, M., Whiteley, W., Mendelow, A.~D., Wardlaw, J.~M., Sudlow, C.,
  and Alex, B. (2022).
\newblock Automated clinical coding: what, why, and where we are?
\newblock {\em npj Digital Medicine}, 5(1):159.

\bibitem[Eisele et~al., 2023]{eisele2023frame}
Eisele, O. et~al. (2023).
\newblock Capturing a news frame: Comparing machine-learning approaches to
  frame analysis with different degrees of supervision.
\newblock {\em Communication Methods and Measures}, 17(3).

\bibitem[Esser et~al., 2018]{esser2018when}
Esser, F. et~al. (2018).
\newblock When communication meets computation: Opportunities, challenges, and
  pitfalls in computational communication science.
\newblock {\em Communication Methods and Measures}, 12(2-3).

\bibitem[Garg et~al., 2022]{garg2022learning}
Garg, A., Sani, D., and Anand, S. (2022).
\newblock Learning hierarchy aware features for reducing mistake severity.
\newblock In {\em European Conference on Computer Vision (ECCV)}, pages
  252--267. Springer.

\bibitem[Grimmer et~al., 2022]{grimmer2022text}
Grimmer, J., Roberts, M.~E., and Stewart, B.~M. (2022).
\newblock {\em Text as Data: A New Framework for Machine Learning and the
  Social Sciences}.
\newblock Princeton University Press.

\bibitem[Grimmer and Stewart, 2013]{grimmer2013text}
Grimmer, J. and Stewart, B.~M. (2013).
\newblock Text as data: The promise and pitfalls of automatic content analysis
  methods for political texts.
\newblock {\em Political Analysis}, 21(3):267--297.

\bibitem[Guo et~al., 2017]{guo2017calibration}
Guo, C., Pleiss, G., Sun, Y., and Weinberger, K.~Q. (2017).
\newblock On calibration of modern neural networks.
\newblock In {\em Proceedings of the 34th International Conference on Machine
  Learning}, volume~70, pages 1321--1330.

\bibitem[Halford et~al., 2005]{halford2005cognitive}
Halford, G.~S., Baker, R., McCredden, J.~E., and Bain, J.~D. (2005).
\newblock How many variables can humans process?
\newblock {\em Psychological Science}, 16(1):70--76.

\bibitem[Hayes and Krippendorff, 2007]{hayes2007reliability}
Hayes, A.~F. and Krippendorff, K. (2007).
\newblock Answering the call for a standard reliability measure for coding
  data.
\newblock {\em Communication Methods and Measures}, 1(1):77--89.

\bibitem[Kiesel et~al., 2019]{kiesel2019semeval}
Kiesel, J., Mestre, M., Shukla, R., Vincent, E., Adineh, P., Corney, D., Stein,
  B., and Potthast, M. (2019).
\newblock Semeval-2019 task 4: Hyperpartisan news detection.
\newblock In {\em Proceedings of the 13th International Workshop on Semantic
  Evaluation}, pages 829--839.

\bibitem[Krippendorff, 2018]{krippendorff2018content}
Krippendorff, K. (2018).
\newblock {\em Content Analysis: An Introduction to Its Methodology}.
\newblock SAGE, 4 edition.

\bibitem[Lakshminarayanan et~al., 2017]{lakshminarayanan2017simple}
Lakshminarayanan, B., Pritzel, A., and Blundell, C. (2017).
\newblock Simple and scalable predictive uncertainty estimation using deep
  ensembles.
\newblock In {\em Advances in Neural Information Processing Systems},
  volume~30, pages 6402--6413.

\bibitem[Ling et~al., 2024]{ling2024domain}
Ling, C., Zhao, X., Wang, J., Tamil, L., et~al. (2024).
\newblock Domain specialization as the key to make large language models
  disruptive: A comprehensive survey.
\newblock {\em arXiv preprint arXiv:2305.18703}.

\bibitem[Lombard et~al., 2002]{lombard2002content}
Lombard, M., Snyder-Duch, J., and Bracken, C.~C. (2002).
\newblock Content analysis in mass communication: Assessment and reporting of
  intercoder reliability.
\newblock {\em Human Communication Research}, 28(4):587--604.

\bibitem[Maier et~al., 2018]{maier2018lda}
Maier, D., Waldherr, A., Miltner, P., et~al. (2018).
\newblock Applying lda topic modeling in communication research: Toward a valid
  and reliable methodology.
\newblock {\em Communication Methods and Measures}, 12(2-3):93--118.

\bibitem[Neuendorf, 2017]{neuendorf2017content}
Neuendorf, K.~A. (2017).
\newblock {\em The Content Analysis Guidebook}.
\newblock SAGE, 2 edition.

\bibitem[Niculescu-Mizil and Caruana, 2005]{niculescu2005predicting}
Niculescu-Mizil, A. and Caruana, R. (2005).
\newblock Predicting good probabilities with supervised learning.
\newblock In {\em Proceedings of the 22nd International Conference on Machine
  Learning}, pages 625--632.

\bibitem[Ouyang et~al., 2024]{liu2023stability}
Ouyang, T., MaungMaung, A.~P., Konishi, K., Seo, Y., and Echizen, I. (2024).
\newblock Stability analysis of chatgpt-based sentiment analysis in ai quality
  assurance.
\newblock {\em Electronics}, 13(24):5043.

\bibitem[Pan and Yang, 2010]{pan2010survey}
Pan, S.~J. and Yang, Q. (2010).
\newblock A survey on transfer learning.
\newblock {\em IEEE Transactions on Knowledge and Data Engineering},
  22(10):1345--1359.

\bibitem[Reason, 1990]{reason1990human}
Reason, J. (1990).
\newblock {\em Human Error}.
\newblock Cambridge University Press, Cambridge.

\bibitem[Ribeiro et~al., 2016]{chen2023quality}
Ribeiro, M.~T., Singh, S., and Guestrin, C. (2016).
\newblock "why should i trust you?": Explaining the predictions of any
  classifier.
\newblock In {\em Proceedings of the 22nd ACM SIGKDD International Conference
  on Knowledge Discovery and Data Mining}, pages 1135--1144.
\newblock Introduces LIME for model interpretability and trustworthiness.

\bibitem[Schwabe et~al., 2024]{zhao2024domain}
Schwabe, D., Houben, S., Schmitz-Koep, B., Brandl, F., Golland, P., Poustka,
  L., Menze, B., Zimmer, C., Sorg, C., and Hedderich, D.~M. (2024).
\newblock The metric-framework for assessing data quality for trustworthy ai in
  medicine: A systematic review.
\newblock {\em npj Digital Medicine}, 7(1):203.
\newblock Systematic framework for data quality assessment in domain-specific
  AI applications.

\bibitem[Soroush et~al., 2024]{soroush2024large}
Soroush, A., Glicksberg, B.~S., Zimlichman, E., Barash, Y., Friedman, C.,
  Tatonetti, N.~P., and Elhadad, N. (2024).
\newblock Large language models are poor medical coders---benchmarking of
  medical code querying.
\newblock {\em NEJM AI}, 1(5).

\bibitem[Trilling and Jonkman, 2018]{trilling2018scaling}
Trilling, D. and Jonkman, J. G.~F. (2018).
\newblock Scaling up content analysis.
\newblock {\em Communication Methods and Measures}, 12(2-3):158--174.

\bibitem[van Atteveldt et~al., 2023]{vanatteveldt2023transfer}
van Atteveldt, W. et~al. (2023).
\newblock Advancing automated content analysis for a new era of media effects
  research: The key role of transfer learning.
\newblock {\em Communication Methods and Measures}.

\bibitem[van Atteveldt et~al., 2021]{vanatteveldt2021validity}
van Atteveldt, W., Huneeus, F., et~al. (2021).
\newblock The validity of sentiment analysis: Comparing manual annotation,
  crowd-coding, dictionary approaches, and machine learning algorithms.
\newblock {\em Communication Methods and Measures}, 15(2):121--140.

\bibitem[Wang et~al., 2023]{wang2023self}
Wang, X., Wei, J., Schuurmans, D., Le, Q., Chi, E., Narang, S., Chowdhery, A.,
  and Zhou, D. (2023).
\newblock Self-consistency improves chain of thought reasoning in language
  models.
\newblock In {\em International Conference on Learning Representations (ICLR)}.

\bibitem[Wei et~al., 2022]{wei2022chain}
Wei, J., Wang, X., Schuurmans, D., Bosma, M., Xia, F., Chi, E., Le, Q.~V.,
  Zhou, D., et~al. (2022).
\newblock Chain-of-thought prompting elicits reasoning in large language
  models.
\newblock In {\em Advances in Neural Information Processing Systems},
  volume~35, pages 24824--24837.

\bibitem[Welbers et~al., 2017]{welbers2017text}
Welbers, K., van Atteveldt, W., and Benoit, K. (2017).
\newblock Text analysis in r for communication research: A tutorial.
\newblock {\em Communication Methods and Measures}, 11(4):245--265.

\bibitem[Widmann and Wich, 2021]{widmann2021three}
Widmann, T. and Wich, M. (2021).
\newblock Three gaps in computational text analysis methods for social
  sciences: A research agenda.
\newblock {\em Communication Methods and Measures}, 16(1):1--18.

\bibitem[Zhang et~al., 2004]{zhang2004cognitive}
Zhang, J., Patel, V.~L., Johnson, T.~R., and Shortliffe, E.~H. (2004).
\newblock A cognitive taxonomy of medical errors.
\newblock {\em Journal of Biomedical Informatics}, 37(3):193--204.

\bibitem[Zhao and Liu, 2025a]{zhao2025complex}
Zhao, Z. and Liu, Y. (2025a).
\newblock Automated quality assessment for llm-based complex qualitative
  coding: A confidence-diversity framework.
\newblock {\em arXiv preprint arXiv:2508.20462}.

\bibitem[Zhao and Liu, 2025b]{zhao2025accessible}
Zhao, Z. and Liu, Y. (2025b).
\newblock A confidence-diversity framework for calibrating ai judgement in
  accessible qualitative coding tasks.
\newblock {\em arXiv preprint arXiv:2508.02029}.

\end{thebibliography}

\end{document}